# Inter-sphere consistency-based method for camera-projector pair calibration

**Zhaoshuai Qi, Jingqi Pang, Yifeng Hao and Yanning Zhang\***

*College of Computer Science, Northwestern Polytechnical University, Xi'an 710072, China,*
*National Engineering Laboratory for Integrated Aero-Space-Ground-Ocean Big Data Application Technology.*
*zhaoshuaiqi1206@163.com*

**Abstract:** We construct constraints from consistency between estimated parameters from different spheres, termed inter-sphere consistency. It facilitates more flexible calibration using only two spheres, which has been considered a challenging and not well addressed ill-posed problem.

## 1. Introdution

Calibration of camera-projector pairs (CPPs) is vitally important for structured-light 3D reconstruction and spatial augmented reality (or projection mapping), etc. Plenty of works have been focused on improving both of the flexibility and accuracy of calibration, between which there is an inherent trade-off. As one of the most successful attempts to balance the above trade-off, Zhang proposed to use only a planar pattern in unknown orientations to calibrate a camera[1] and achieved flexible and accurate calibration, which was rapidly adapted to CPP calibration. After that, as the most competitive methods, sphere-based methods calibrated camera and projector only from images of spheres instead of planar pattern, achieving more flexibility with comparable accuracy. Since an image of two spheres provides two independent constraints for the internal parameter of camera[2, 3], at least three spheres are required for full calibration. Continuing to reduce the number of spheres, say only two spheres, will gain more flexibility and further simplify the calibration for generic usage, which, however, makes the calibration a much more challenging and ill-posed problem due to insufficient constraints.

In the paper, we consider the above ill-posed calibration with only two spheres. It is shown that rather than independent constraints from two distinctive spheres, additional constraints can be constructed from strong correlation between these spheres, which has never been well exploited before. More specifically, estimated CPP parameters exhibit significant consistency across different spheres, termed inter-sphere consistency(ISC), which provides sufficient constraints to ease the ill-posedness and facilitate an accurate calibration.

## 2. Fundamental principle

*2.1* Correlation between $\mathbf{M}_P$ and $\mathbf{K}_C$ with respect to a single sphere

Place a sphere with a known radius of r in the scene, and regard the projector as a "reverse camera". The projector model can be expressed as:

$$\lambda_P \begin{bmatrix} x_P \\ y_P \\ 1 \end{bmatrix} = \mathbf{K}_P \left[ \mathbf{R} \mid T \right] X = \mathbf{M}_P X, \qquad (1)$$

where, $x_{Pi} = (x_{Pi}, y_{Pi})$ is the projector image coordinates, $\mathbf{R}$ and $T$ is the projector's rotation matrix and translation vector relative to the camera. $X$ is the world coordinate of the point on the sphere, and $\mathbf{M}_P$ is the projector parameter matrix.

With dense coding strategy, say phase-shifting coding with vertical and horizontal fringe patterns, we can find the corresponding projector image coordinate $x_{Pi}$ of camera image coordinates $x_{Ci} = (x_{Ci}, y_{Ci})$ as $x_{Pi} = \varphi_V(x_{Ci}, y_{Ci})/2\pi f_V$, $y_{Pi} = \varphi_H(x_{Ci}, y_{Ci})/2\pi f_H$, where $\varphi_V$, $\varphi_H$ and $f_V, f_H$ are phase maps and frequencies of vertical and horizontal fringe patterns, respectively.

Given $\mathbf{K}_C$ and r, the spherical center coordinate $X_S$ can be easily estimated from the contour image $\mathbf{C}$ of the sphere through singular value decomposition. The world coordinate $X$ of the point on the sphere can be estimated from $X_S$ and $\mathbf{K}_C$ according to the constraint relationship between the sphere center and the spherical coordinate point and the camera imaging model. In this way, we get $x_{Pi}$ and $X$, and substitute them into formula 1, we can easily estimate $\mathbf{M}_P$.

*2.2* Derivation of ISC

Provided an image of two spheres and the same $\mathbf{K}_C$, two $\mathbf{M}_P$'s will be estimated from these spheres, say $\mathbf{M}_{P1}$ and $\mathbf{M}_{P2}$, which should share high consistency, namely ISC, with each other. A best estimate of $\mathbf{K}_C$ should yield a $\mathbf{M}_P$ satisfying such ISC assumption. Combined with the orthogonal constraints , we construct an optimization framework to search for the optimal $\mathbf{K}_C$.

$$\mathbf{K}_C^* = \arg\min \sum_{i=1}^{N_1}\|\mathbf{x}_{1Pi} - \mathbf{M}_P \mathbf{X}_{1i}\| + \sum_{j=1}^{N_2}\|\mathbf{x}_{2Pj} - \mathbf{M}_P \mathbf{X}_{2j}\| \text{ subject to } \mathbf{l} = \mathbf{K}_C^{-T}\mathbf{K}_C^{-1}\mathbf{v}, \qquad (2)$$

where $\mathbf{K}^*_C$ is the optimal $\mathbf{K}_C$ best fitting the ISC, $\{\mathbf{x}_{1Pi}, \mathbf{X}_{1i}\}$, $i=1,2,…,N_1$, and $\{\mathbf{x}_{2Pj}, \mathbf{X}_{2j}\}$, $j=1,2,…,N_2$ are pairs of projector image coordinates and 3D world coordinates of points on two spheres, where the vanishing line $\mathbf{l}$ is one of the eigenvector of $\mathbf{C}_2\mathbf{C}_1^*$, and vanishing point $\mathbf{v}$ is the cross product of the other two eigenvectors, conics $\mathbf{C}_1$ and $\mathbf{C}_2$ are an image of two spheres' contour.

The best result of $\mathbf{K}_C$ can better satisfy the ISC hypothesis, so as to find the best $\mathbf{M}_P$ according to the principle of correlation between $\mathbf{M}_P$ and $\mathbf{K}_C$ with respect to a single sphere. Above all, we achieve a flexible and fast structured light calibration.

## 3. Result

To validate the proposed ISC technique, extensive experiments were conducted. Two CPPs, denoted CPP A and CPP B respectively, with different parameters were implemented, where the former consists of a high-resolution camera with 3384×2704 pixels and a projector with 854×480 pixels and the latter is of the same projector and a low-resolution camera of 1920×1200 pixels.

Performances were compared against the plane-based calibration(PBC)[4] and the most recently reported sphere-based calibration(SBC)[5], respectively. PBC used a calibration board translated by a linear stage with high positioning precision, of which the results were treated as the ground truth. SBC used three different spheres to present in the image, which was the baseline method. Calibration results are shown in Table 1. The errors of most parameters estimated by our method for CPP *A* are less than 10%, indicating an accurate calibration. Reconstruction results of our method ISC is visually identical to PBC and SBC. As shown in Fig. 1, it implies a comparably good performance as SBC. Please note that values in columns labeled "SBC" and "Ours" are relative errors in percentage(%).

Table 1. Calibration results for different CPPs

| Unit/pixel | CPP A | | | | | | CPP B | | | | | |
|---|---|---|---|---|---|---|---|---|---|---|---|---|
| | Camera (3384×2704) | | | Projector (854×480) | | | Camera (1920×1200) | | | Projector (854×480) | | |
| | PBC | SBC | Ours | PBC | SBC | Ours | PBC | SBC | Ours | PBC | SBC | Ours |
| $f_x$ | 3277.5 | 8.1 | 7.5 | 1202.7 | 4.6 | 6.9 | 1791.1 | 19.5 | 18.8 | 1202.7 | 21.4 | 18.4 |
| $f_y$ | 3277.8 | 8.2 | 7.9 | 1199.0 | 5.3 | 7.8 | 1789.2 | 19.9 | 24.4 | 1199.0 | 21.8 | 24.1 |
| $\gamma$ | -18.6 | -0.9 | -0.4 | -8.2 | -2.4 | -2.8 | -1.4 | 1.5 | -54.9 | -8.2 | -0.6 | 3.8 |
| $u_0$ | 1699.4 | 0.3 | 0.7 | 390.7 | -4.7 | -9.5 | 944.9 | 2.8 | 7.1 | 390.7 | -11.0 | 0.9 |
| $v_0$ | 1330.1 | 2.6 | 2.9 | 222.8 | -10.3 | -11.7 | 561.4 | 3.2 | 1.4 | 222.8 | -2.2 | 13.2 |

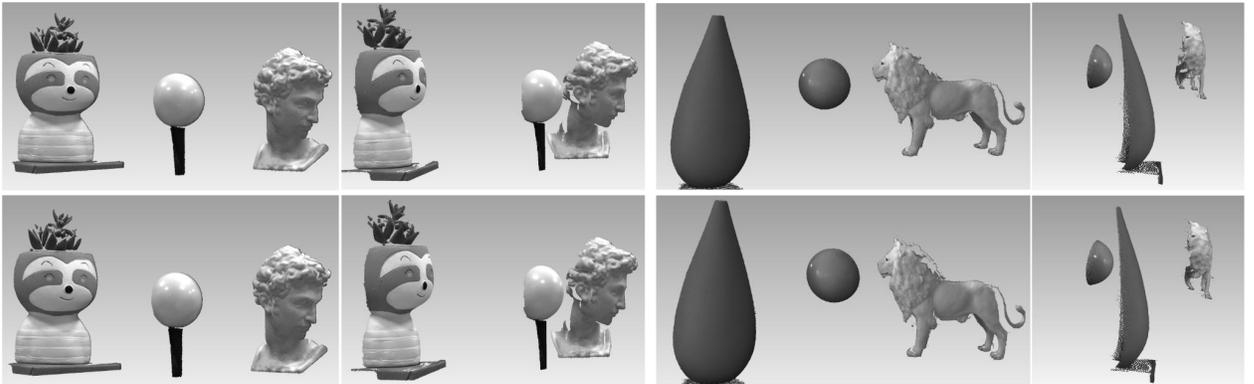

Fig. 1. (*from top to bottom*) The two scenes reconstrued (front view and side view) from CPP *A* and *B* using SBC and our method.

## 4. Conclusion

An inter-sphere consistency-based CPP calibration method has been proposed. Rather than exploiting independent constrains from individual spheres, we construct additional constraints of ISC based on the strong correlation between spheres. The proposed ISC, which has never been well exploited before, provides sufficient constraints and yields an accurate calibration from images of only two spheres, which is known as an ill-posed problem for state-of-the-arts sphere-based methods. Extensive experimental results verified the effectiveness of the proposed method, which demonstrates comparable calibration and reconstruction accuracy with less spheres than state of the arts.

## 5. References

[1] Z. Zhang, "A flexible new technique for camera calibration," TPAMI **22,** 1330-1334 (2000).
[2] M. Agrawal and L. S Davis, "Camera calibration using spheres: A semi-definite programming approach", in ICCV of Proceeding,( International Conference on Computer Vision, Bei Jing 2003),pp. 782–782.
[3] H. Zhang, K. Wong and G. Zhang, "Camera calibration from images of spheres", TPAMI **29,** 499–502 (2007).
[4] S. Zhang and Peisen S Huang, "Novel method for structured light system calibration",OPT ENG **45,** (2005).
[5] X. Yin, M. Ren and L. Zhu, "A single pose series sphere-based calibration method for camera projector structured light system", OPT COMMUN **507,** (2021).